\documentclass[letterpaper, 10pt, journal, twoside]{IEEEtran}

\usepackage{cite}
\usepackage{amsmath, amssymb}
\usepackage{graphicx}
\usepackage{booktabs}
\usepackage{multirow}
\usepackage{siunitx}
\usepackage{xcolor}
\usepackage{xspace}
\usepackage{tikz}
\usetikzlibrary{arrows.meta, decorations.pathreplacing, patterns, positioning, calc}
\usepackage[hidelinks]{hyperref}
\hypersetup{pdftitle={LuxTac: A Flexible, Scalable, and Open-Source Optical Taxel Array with Distributed Three-Axis Force Sensing for Tactile Robot Learning}, pdfauthor={Md Rakibul Islam Prince, John Pelan, Zhiyuan Zhang, Adyaa Khera, Katherine Pesetski, Adeesh Desai, Akimizu Sugiyama, Minwoo Ki, Benjamin Alan Capelle, Temitope Ibrahim Amosa, Zijie Xu, Yihan Huang, Taehun Chung, Long Wang, Wei Li, Shaoting Lin, Yu She}}

\newcommand{\sensor}{LuxTac\xspace}

\makeatletter
\newcommand{\figcaption}{\def\@captype{figure}\caption}
\makeatother

\newcommand{\raw}{x}
\newcommand{\nsig}{s}
\newcommand{\dep}{\delta}

\begin{document}
\raggedbottom

\title{\sensor: A Flexible, Scalable, and Open-Source Optical Taxel Array with Distributed Three-Axis Force Sensing for Tactile Robot Learning}

\author{Md~Rakibul~Islam~Prince$^{1}$, John~Pelan$^{1}$, Zhiyuan~Zhang$^{2}$, Adyaa~Khera$^{3}$, Katherine~Pesetski$^{1}$, Adeesh~Desai$^{2}$, Akimizu~Sugiyama$^{4}$, Minwoo~Ki$^{1}$, Benjamin~Alan~Capelle$^{3}$, Temitope~Ibrahim~Amosa$^{2}$, Zijie~Xu$^{5}$, Yihan~Huang$^{6}$, Taehun Chung$^{7}$, Long~Wang$^{6}$, Wei~Li$^{5}$, Shaoting~Lin$^{7}$, and~Yu~She$^{1, 2, 4}$%
\thanks{This work was supported in part by the Department of the Air Force under Contract FA857126C0002; in part by the National Science Foundation under Awards 2423068 and 2520136; in part by the U.S. Department of Agriculture under Awards 2023-67021-39072 and 2024-67021-42878; and in part by the Showalter Trust.}%

\thanks{$^{1}$Elmore Family School of Electrical and Computer Engineering, Purdue University, West Lafayette, IN 47907, USA. \texttt{\{prince26, jpelan, kpesetski, ki0\}@purdue.edu}}%
\thanks{$^{2}$Edwardson School of Industrial Engineering, Purdue University, West Lafayette, IN 47907, USA. \texttt{\{zhan5570, desai274, tamosa, shey\}@purdue.edu}}%
\thanks{$^{3}$Department of Robotics Engineering Technology, Purdue University, West Lafayette, IN 47907, USA. \texttt{\{khera2, bcapelle\}@purdue.edu}}%
\thanks{$^{4}$Department of Mechanical Engineering, Purdue University, West Lafayette, IN 47907, USA. \texttt{asugiyam@purdue.edu}}%
\thanks{$^{5}$Department of Civil Engineering, Stony Brook University, Stony Brook, NY 11794, USA. \texttt{\{zijie.xu, wei.li.6\}@stonybrook.edu}}%
\thanks{$^{6}$Department of Mechanical Engineering, Stevens Institute of Technology, Hoboken, NJ 07030, USA. \texttt{\{yhuang101, lwang4\}@stevens.edu}}%
\thanks{$^{7}$Department of Mechanical Engineering, University of Wisconsin-Madison, Madison, WI 53706, USA. \texttt{\{tchung29, slin88\}@wisc.edu}}%
}
\IEEEaftertitletext{%
  \vspace{-0.8\baselineskip}%
  \centering
  \noindent\makebox[\textwidth][c]{\includegraphics[width=\textwidth]{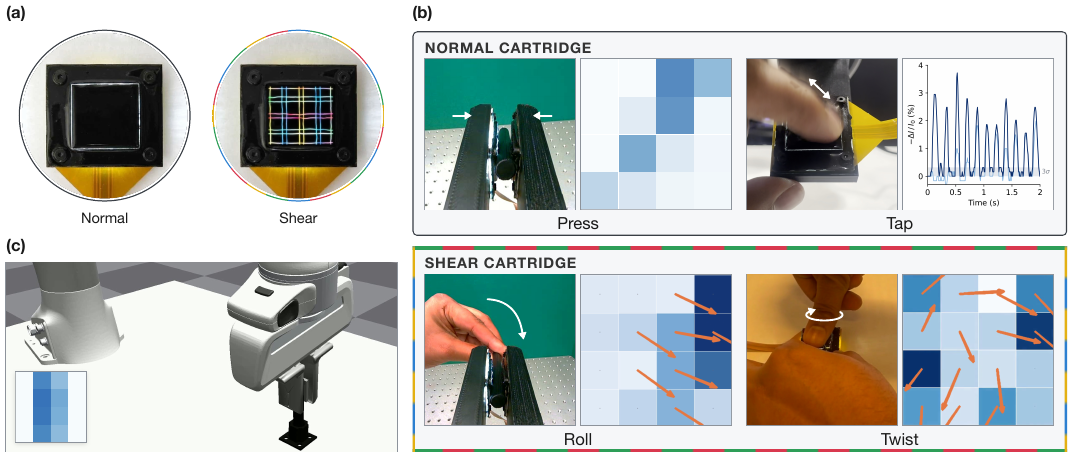}}\par
   \figcaption{\textbf{\sensor overview.} (a)~The normal and shear gel cartridges, which mount on the same taxel board. (b)~Real-world interactions (white arrows indicate the direction of motion) and the corresponding \sensor readings. With the normal cartridge, pressing yields a per-taxel pressure map, and tapping produces repeatable 2--4\% intensity drops above the $3\sigma$ noise floor. With the shear cartridge, rolling and twisting yield per-taxel pressure (blue) and shear (orange arrows). (c)~Simulated grasp in IsaacGym with TacSL, with the simulated tactile reading (inset).}
  \label{fig:overview}%
  \vspace{0.8\baselineskip}%
}

\maketitle

\begin{abstract}
Low-cost tactile arrays have made touch a practical input for robot learning, but piezoresistive designs sense only pressure, while camera-based sensors that also resolve shear are too bulky to tile over a gripper. To solve this problem, we present \sensor, a scalable optical taxel array, about \SI{3.5}{\milli\meter} thick and built from off-the-shelf parts. Each taxel is a digital RGBC light-to-digital converter under a translucent white silicone layer capped by an opaque black film, so the back-scattered LED light falls monotonically with \emph{local indentation depth}. The same board accepts two interchangeable gel cartridges: a plain one for normal sensing, and one with colored walls that encodes the local shear in the color of the returned light, with no camera, magnets, or bump overlay, making \sensor, to our knowledge, the first optical taxel array to resolve shear at individual taxels. Characterized with an automated CNC indenter and a six-axis force/torque sensor, the normal cartridge responds linearly to force ($R^2 = 0.999$) from \SI{104}{\milli\newton} to at least \SI{5.8}{\newton}, with 2.0\% repeatability, 5.9\% hysteresis, and a taxel-to-taxel variation of 6--10\% after normalization. With the shear cartridge, per-taxel models estimate the three-axis force at every taxel; across three indenter shapes and nine held-out sessions, the summed estimates reach $R^2 = 0.96/0.96/0.92$ in $x/y/z$, each taxel's estimate $0.92/0.93/0.94$, and the median shear-direction error is \SI{4.8}{\degree}. Adding \sensor pressure readings to an Action Chunking Transformer policy substantially improves peg-in-hole success over a vision-only baseline. Design files, code, and videos are available at \url{https://luxtac.github.io}.
\end{abstract}


\section{Introduction}
\IEEEPARstart{T}{ouch} tells a robot what vision cannot: whether it is holding an object, how hard, and where the contact is when the fingers occlude the scene~\cite{dahiya2010tactile}. Even low-cost tactile arrays substantially improve learned manipulation policies when fused with vision~\cite{huang2024vitac,huang2026flexitac,slepyan2026opentac}. The most widely adopted are piezoresistive: a force-sensitive film between orthogonal electrodes, read as a resistance matrix~\cite{sundaram2019stag,huang2024vitac,huang2026flexitac}. They are thin, flexible, and cheap, and their open-source releases have lowered the barrier to tactile learning considerably.

Piezoresistive arrays have two weaknesses that matter once a sensor leaves the bench. Each taxel reports a single number, so they sense pressure but not shear. And that number is a contact resistance through a polymer composite, which drifts, saturates early, depends on the loading history, and differs between pads.

Optical transduction offers a different trade-off. Camera-based sensors such as GelSight~\cite{yuan2017gelsight}, DIGIT~\cite{lambeta2020digit}, and TacTip~\cite{wardcherrier2018tactip} reconstruct the deformation field of an elastomer at high resolution, at the cost of bulk, bandwidth, and high-dimensional observations. Miniaturized designs such as TacScope~\cite{prince2025tacscope} and MVP-Tac~\cite{prince2026mvp} shrink the sensing surface, but the camera's optical path still makes them thick along the viewing axis. Between these extremes, discrete LED--photodetector pairs embedded in elastomer have been used to sense contact and proximity~\cite{patel2016proximity,patel2018proximity,piacenza2020finger}. Such optical taxels measure a geometric quantity, the deformation of the elastomer above them, through a stable, linear photodetector with an on-chip digital converter.

\begin{figure}[t]
  \centering
  \includegraphics[width=\linewidth]{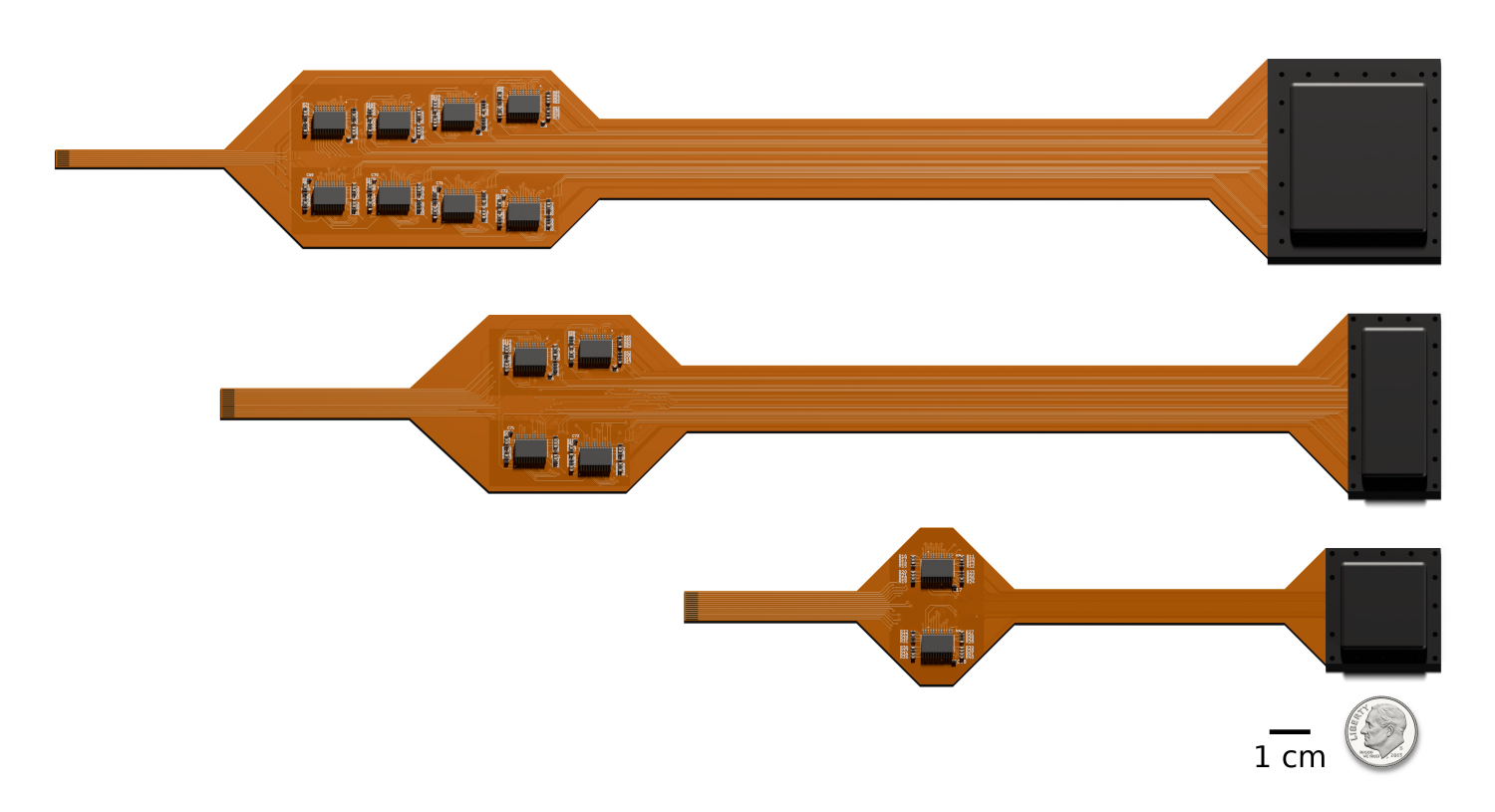}
    \caption{\textbf{Scalability of \sensor (renderings)}. Each taxel is an independent digital detector on its own bus segment, therefore, the same taxel cell tiles into arrays of other sizes and aspect ratios without changing the readout scheme. Top to bottom: $8{\times}8$ and $8{\times}4$ arrays, and the prototype board. A US dime (\SI{17.9}{\milli\meter}) and a \SI{1}{\centi\meter} bar are shown for scale.}
  \label{fig:scale}
\end{figure}

We develop this idea into \sensor (Fig.~\ref{fig:overview}), a flexible, dense, and scalable tactile array. Each taxel is a commercial light-to-digital converter~\cite{ams_tcs3472} under a translucent silicone layer and an opaque black film; compressing the silicone moves the absorbing film closer to the detector and reduces the returned light. Because every taxel is an independent digital device on its own bus segment, the same taxel cell tiles into arrays of any size and aspect ratio without changing the readout (Fig.~\ref{fig:scale}). Because each detector also reports color, a second gel cartridge with colored walls adds shear sensing at every taxel without changing the electronics, which makes \textbf{\sensor, to our knowledge, the first optical taxel array that resolves shear per taxel.} Our contributions are:
\begin{enumerate}
  \item An approximately \SI{3.5}{\milli\meter} thick optical taxel array built entirely from off-the-shelf parts on a four-layer flexible PCB, in which every taxel is an independent digital RGBC detector on its own I$^2$C bus segment, with no analog front end, so the same cell tiles into arrays of any size by adding multiplexers and buses (Sec.~\ref{sec:design}).
  \item Two interchangeable gel cartridges: the shear cartridge encodes the local shear above each taxel in the color of the returned light, so the same board senses either normal deformation, or normal deformation and shear, at every taxel (Secs.~\ref{sec:elastomer} and~\ref{sec:shear_results}).
  \item Characterization over 240 normal presses and shear tests with three indenter shapes in nine sessions, evaluated at contact positions and in sessions held out from training; sim-to-real alignment of contact patterns; and two real-world tasks, peg-in-hole insertion and cube sorting, with an Action Chunking Transformer policy (Secs.~\ref{sec:results}--\ref{sec:robot}).
\end{enumerate}

\section{Related Work}

Piezoresistive textiles and laminates are the workhorse of large-area tactile sensing~\cite{sundaram2019stag}. 3D-ViTac~\cite{huang2024vitac} built $16{\times}16$ Velostat pads for bimanual grippers and fused their readings with vision as a 3D point cloud for diffusion policies~\cite{chi2023diffusion,qi2017pointnetpp}. FlexiTac~\cite{huang2026flexitac} replaced hand-laid conductive thread with FPC electrodes, raised the readout to \SI{100}{\hertz}, and added GPU tactile simulation for real-to-sim-to-real fine-tuning. OpenTac~\cite{slepyan2026opentac} generates resistive skins automatically from robot geometry and adds a bump overlay that converts shear into differential pressure across several taxels. Magnetic skins such as ReSkin and AnySkin~\cite{bhirangi2021reskin,bhirangi2024anyskin} offer replaceable three-axis sensing, and uSkin~\cite{tomo2018uskin} reports three-axis force at each taxel from magnets above Hall sensors, at the cost of sensitivity to external magnetic fields. SpikeATac~\cite{chang2025spikeatac} pairs a \SI{4}{\kilo\hertz} taxelized PVDF array with capacitive pads to capture contact onset. \sensor keeps the thin, flexible, tileable form of these arrays but replaces the resistive or magnetic transducer with an optical one: each taxel reports the local indentation of the elastomer above it, and the shear cartridge reads shear optically, from color, rather than from magnets or bump patterns.

Vision-based sensors~\cite{yuan2017gelsight,lambeta2020digit,wardcherrier2018tactip, prince2026mvp} image the inside of an elastomer and recover contact geometry at sub-millimeter resolution, but need a camera and optical path that make them thick and difficult to tile. Several groups instead embed discrete emitter--detector pairs: Patel and Correll used commodity proximity sensors under elastomer to measure distance and contact force~\cite{patel2016proximity,patel2018proximity}, and Piacenza~\emph{et al.} learned contact location and force from overlapping LED--photodiode signals in a curved finger~\cite{piacenza2020finger}. SpectraTac~\cite{wu2026spectratac} placed three RGBC color sensors and RGB LEDs under a transparent elastomer with a black shielding layer and regressed one 3D force and contact region for the whole sensor from its 12 color channels with a neural network. These works rely on learned models of coupled channels. \sensor uses a regular grid of independent, digitally read taxels under an absorbing top layer that blocks ambient light, so each taxel is read and normalized individually, maps one-to-one onto a simulated taxel, and reports its own shear.

Simulators for camera-based sensors render the elastomer image from contact geometry~\cite{wang2022tacto,si2022taxim,akinola2024tacsl}. For taxel arrays, FlexiTac converts the penetration of each simulated taxel into force with a Kelvin--Voigt penalty model whose stiffness and damping are tuned until the simulated force response matches the real pad~\cite{huang2026flexitac}; OpenTac pairs a MuJoCo sensor model with per-sensor calibration~\cite{slepyan2026opentac}. Because a \sensor taxel responds to local indentation, our simulated taxel maps its penetration depth directly to the normalized signal using the depth response measured on the real pad, and we compare the resulting contact patterns with real grasps (Sec.~\ref{sec:sim}).

\section{Sensor Design}\label{sec:design}

\subsection{Sensing Principle}
\begin{figure}[t]
    \centering
    \includegraphics[width=\linewidth]{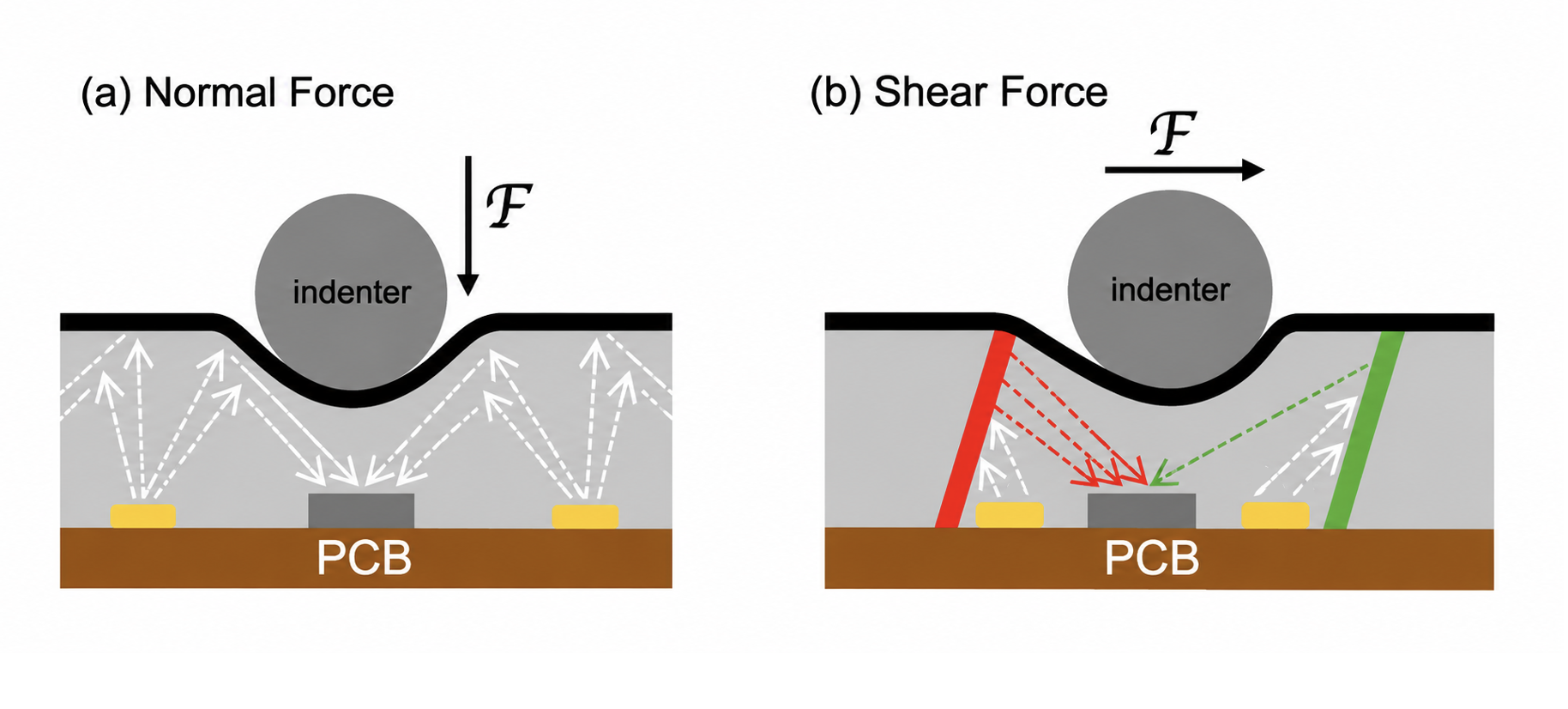}
    \caption{\textbf{Sensing principle under normal and shear loading.}
    (a) Normal loading produces symmetric deformation,
    altering the optical paths between the LEDs and photodetector.
    (b) Shear loading produces asymmetric deformation,
    changing the light propagation and intensity received
    by the photodetector.}
    \label{fig:principle}
\end{figure}

Each taxel consists of a photodetector and nearby LEDs under a layer of translucent white silicone of thickness $d = \SI{2}{\milli\meter}$, capped by an opaque black film (Fig.~\ref{fig:principle}). LED light entering the silicone is scattered by the white pigment; part of it returns to the detector, and part reaches the black film and is absorbed. The black film also shields the detector from ambient light. In the shear cartridge, colored silicone walls divide the white layer into one cube per taxel: the walls between neighboring columns alternate green and red, and the walls between neighboring rows alternate yellow and blue (Sec.~\ref{sec:elastomer}), so each taxel is bordered by walls of different colors. Shear tilts these walls and changes how much light of each color reaches the detector below: shear along the rows ($x$, across the column walls) shifts the red--green balance, and shear along the columns ($y$, across the row walls) the yellow--blue balance. Brightness therefore encodes normal deformation, and chromaticity encodes the direction and magnitude of shear (Sec.~\ref{sec:shear_results}).

\subsection{Electronics}

\begin{figure*}[!t]
    \centering
    \includegraphics[
        width=0.95\textwidth,
        height=0.42\textheight,
        keepaspectratio
    ]{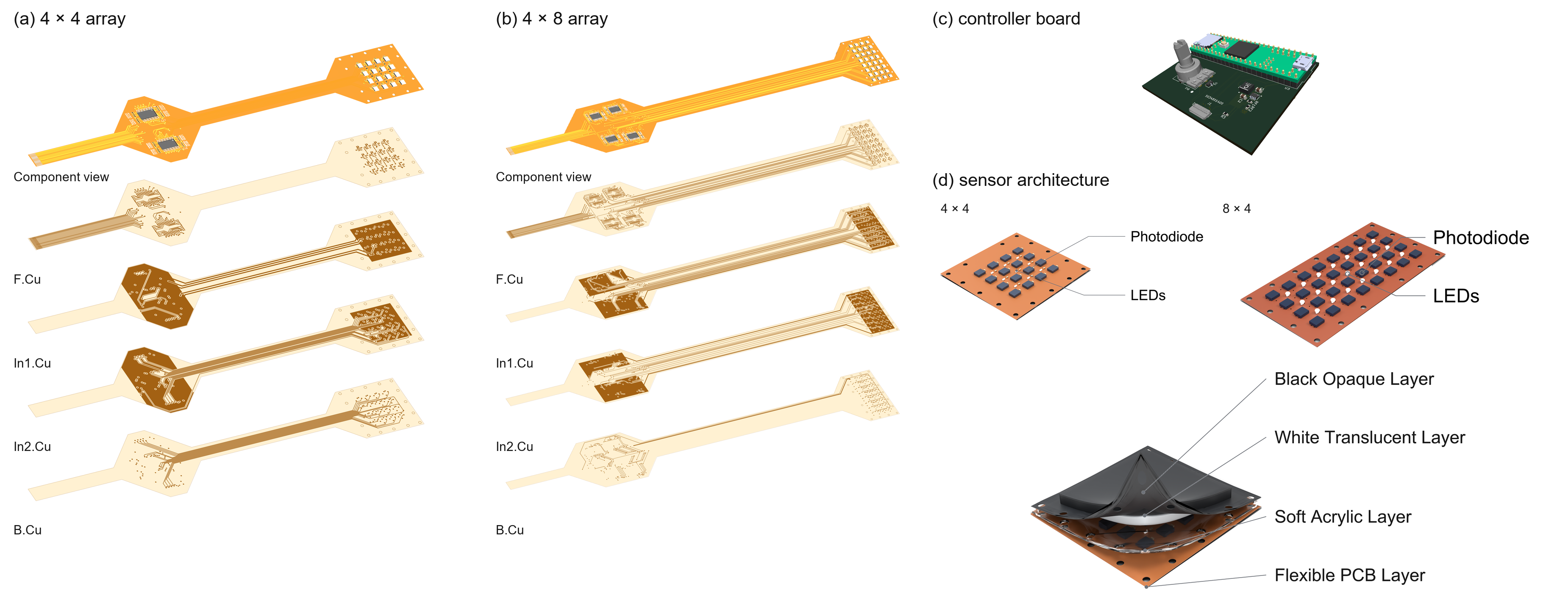}
    \caption{\textbf{Electronics architecture of \sensor.}
    (a) Exploded isometric view of the prototype's
    flexible printed circuit board (FPCB) with 16 RGBC light-to-digital converters.
    (b) Exploded isometric view of the $4{\times}8$
    FPCB with 32 RGBC light-to-digital converters.
    Both designs show the component assembly and
    four copper layers (F.Cu, In1.Cu, In2.Cu, and B.Cu).
    (c) Controller and data-acquisition PCB containing
    the Teensy~4.1, power regulation, and sensor
    interfacing circuitry.
    (d) Sensing-head component arrangement and
    layered tactile sensor construction.}
    \label{fig:electronics}
\end{figure*}

The sensing head is a \SI{21.8}{} $\times$ \SI{23.8}{\milli\meter} region of a four-layer, \SI{0.2}{\milli\meter} thick flexible printed circuit board (FPCB) (Fig.~\ref{fig:electronics}). The prototype characterized in this paper carries 16 TCS34725 RGBC light-to-digital converters~\cite{ams_tcs3472} on a rectangular grid with pitches of \SI{4.42}{\milli\meter} (columns) and \SI{4.80}{\milli\meter} (rows), and nine 0603 chip-scale LEDs arranged in a $3{\times}3$ grid, with each LED centered between four adjacent photodiodes. Each converter integrates an IR-blocking filter, a photodiode array, and a 16-bit integrating ADC per channel, so the analog signal never leaves the package. The converters operate at the shortest integration time of \SI{2.4}{\milli\second} and $4\times$ analog gain~\cite{ams_tcs3472}.

The same electronics serve both gel cartridges: with the normal cartridge only the clear (C) channel is read, and with the shear cartridge all four red, green, blue, and clear (RGBC) channels.

All TCS34725 devices share one fixed I$^2$C address, so the head is connected through a narrow neck to two TCA9548A eight-channel I$^2$C switches~\cite{ti_tca9548a}, giving each sensor its own bus segment with dedicated \SI{4.7}{\kilo\ohm} pull-up resistors. The switches communicate with a Teensy~4.1 through two independent I$^2$C buses. A single NMOS transistor switch gates all nine LEDs, which allows LED-off frames for ambient-light checks. A Teensy~4.1 polls the sensors and streams frames containing either one clear-channel reading or one set of RGBC readings per taxel to the host over USB serial at \SI{100}{\hertz} in this work;  the I$^2$C timing of an RGBC scan of all taxels (about \SI{5.5}{\milli\second}) allows frame rates up to about \SI{180}{\hertz} (Table~\ref{tab:compare}). The complete FPCB pad and PCB readout board are shown in Fig.~\ref{fig:electronics}.

\begin{table}[!htb]
\caption{Comparison with representative tactile arrays}
\label{tab:compare}
\centering
\footnotesize
\setlength{\tabcolsep}{3pt}
\begin{tabular}{@{}lllll@{}}
\toprule
Sensor & Transduction & Shear/taxel & Rate & Sim. \\
\midrule
3D-ViTac~\cite{huang2024vitac} & Piezoresistive & -- & \SI{32}{\hertz} & -- \\
uSkin~\cite{tomo2018uskin} & Magnetic (Hall) & \checkmark & \SI{100}{\hertz} & -- \\
FlexiTac~\cite{huang2026flexitac} & Piezoresistive & -- & \SI{100}{\hertz} & \checkmark \\
OpenTac~\cite{slepyan2026opentac} & Piezores.\ + bumps & --$^{*}$ & \SI{67}{\hertz}$^\dagger$ & \checkmark \\
SpikeATac~\cite{chang2025spikeatac} & PVDF + capacitive & -- & \SI{4}{\kilo\hertz} / \SI{40}{\hertz} & -- \\
\textbf{\sensor (ours)} & \textbf{Optical, digital} & \checkmark & \textbf{100--180\,Hz}$^\S$ & \checkmark \\
\bottomrule
\end{tabular}\\[2pt]
\raggedright\scriptsize $^{*}$Shear is inferred from the pressure pattern of several taxels under each bump, not measured by each taxel. $^\dagger$Full raster scan of a 1024-taxel skin; adaptive partial sampling reaches \SI{1286}{\hertz}~\cite{slepyan2026opentac}. $^\S$Streamed at \SI{100}{\hertz} in this work; up to \SI{180}{\hertz} from the bus scan time.

\end{table}

Table~\ref{tab:bom} lists the bill of materials. The pad costs \$100.7 and the readout board \$37.5. Since 97\% of the pad cost is the light-to-digital converters, the cost grows linearly with the number of taxels, at about \$6.3 per taxel. The converter therefore sets the cost of the array, and a next-generation board built around a newer, lower-cost color sensor is in development, which reduces the per-taxel cost several-fold.

\begin{table}[t]
\caption{Bill of materials of the prototype (unit prices in USD)}
\label{tab:bom}
\centering
\footnotesize
\setlength{\tabcolsep}{4pt}
\begin{tabular*}{\columnwidth}{@{}l@{\extracolsep{\fill}}rrr@{}}
\toprule
Part & Qty & Unit & Total \\
\midrule
\multicolumn{4}{@{}l}{\textbf{Sensing pad}} \\
\quad TCS34725FN RGBC sensor & 16 & 6.121 & 97.94 \\
\quad White SMD LED & 9 & 0.071 & 0.64 \\
\quad TCA9548APWR I$^2$C switch & 2 & 0.822 & 1.64 \\
\quad Passives (R, C)$^{a}$ & 59 & -- & 0.52 \\
\quad \textit{Pad total} & & & \textit{100.74} \\
\midrule
\multicolumn{4}{@{}l}{\textbf{Readout board}} \\
\quad Teensy~4.1 & 1 & 31.50 & 31.50 \\
\quad MP2161GJ-Z buck converter & 1 & 1.822 & 1.82 \\
\quad \SI{1}{\micro\henry} inductor & 1 & 1.940 & 1.94 \\
\quad \SI{10}{\kilo\ohm} potentiometer & 1 & 1.780 & 1.78 \\
\quad FPC connector & 1 & 0.309 & 0.31 \\
\quad NMOS, resistors, capacitors$^{b}$ & 10 & -- & 0.17 \\
\quad \textit{Board total} & & & \textit{37.52} \\
\bottomrule
\end{tabular*}\\[2pt]
\parbox{0.95\linewidth}{\scriptsize $^{a}$18$\times$\SI{100}{\nano\farad}, 9$\times$\SI{68}{\ohm}, 32$\times$\SI{4.7}{\kilo\ohm}. $^{b}$1 NMOS; \SI{200}{\kilo\ohm}, \SI{44.2}{\kilo\ohm}, \SI{10}{\kilo\ohm}, 4$\times$\SI{4.7}{\kilo\ohm}; 2$\times$\SI{10}{\micro\farad}.}
\end{table}

\subsection{Elastomer Stack and Fabrication}\label{sec:elastomer}

\begin{figure}[!tp]
    \centering
    \includegraphics[
        width=\columnwidth,
        keepaspectratio
    ]{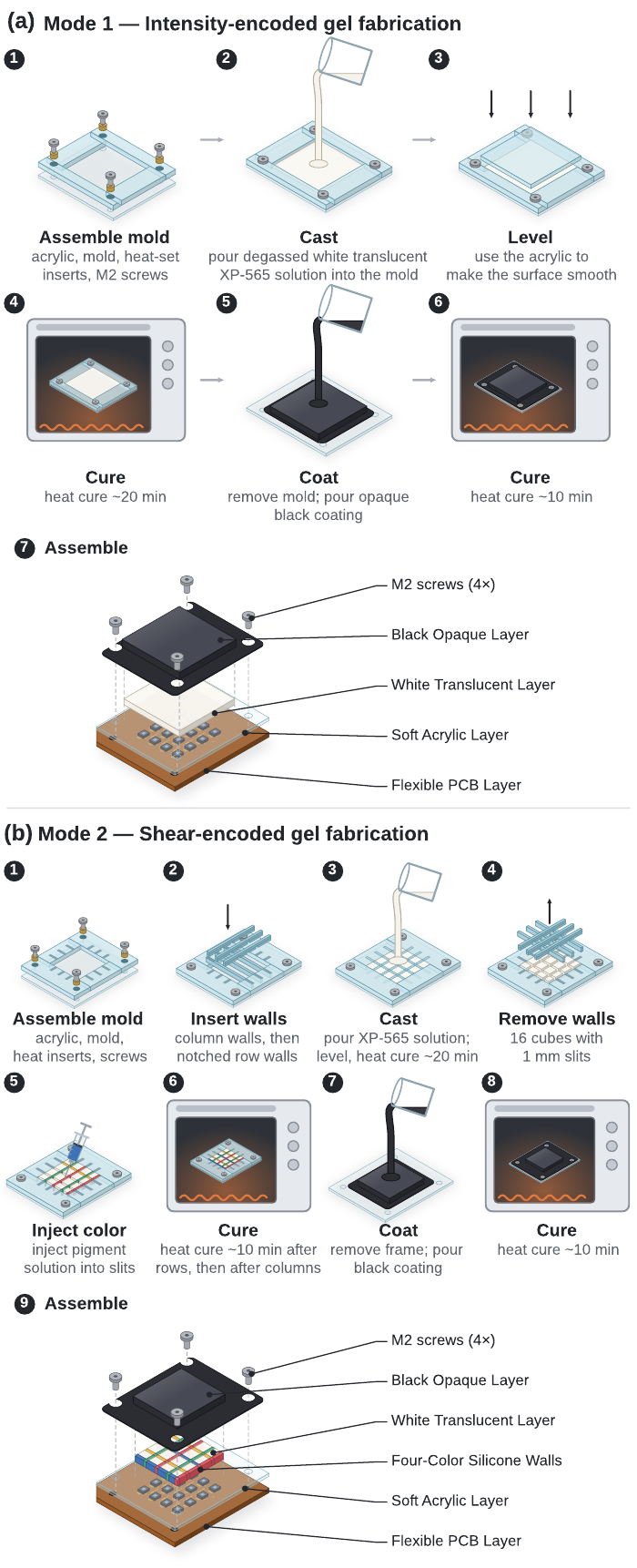}
\caption{\textbf{\sensor Fabrication.} (a)~Mode~1 (intensity-encoded), with a white translucent layer under a
black opaque layer, and (b)~Mode~2 (shear-encoded), with four-color silicone walls
between white translucent cubes, one per taxel. Each gel is mold-cast and heat-cured on a soft
acrylic layer, coated with the black opaque layer, and mounted on the FPCB
layer above the photodiode array.}
    \label{fig:fabrication}
\end{figure}

The elastomer stack is made in two variants with the same materials (Fig.~\ref{fig:fabrication}): Mode~1 (intensity-encoded) has a continuous white layer under the black film, and Mode~2 (shear-encoded) divides it into one cube per taxel, separated by colored silicone walls.

The white translucent layer is prepared from XP-565 silicone mixed at a 15:1 ratio by weight and pigmented with 0.04\,wt\% white Smooth-On Silc Pig. This low pigment concentration keeps the layer translucent, so light can pass through it and reach the overlying black film, preserving the sensor's sensitivity to changes in the film-to-sensor distance. After mixing, the silicone is degassed under vacuum until no visible bubbles remain. The black opaque layer is prepared by mixing Smooth-On Psycho Paint Part~A, Part~B, black Silc Pig, and NOVOCS silicone solvent at a 1:1:0.5:1 ratio by weight and the color pigments use the same recipe where the black Silc Pig is replaced by the respective color pigment (red, green, yellow or blue).

Both variants are cast in laser-cut acrylic molds as shown in Fig.~\ref{fig:fabrication} (a)(1) and Fig.~\ref{fig:fabrication} (b)(1). Each mold consists of a soft, flexible, clear acrylic base plate (0.5~mm), which remains part of the final stack, and a 2~mm acrylic frame fastened to the base with four M2 screws. The base plate is primed before casting so that the silicone bonds to it. The Mode~1 frame defines a $21.5\times21.5$~mm cavity; the Mode~2 frame defines a $16.4\times18.0$~mm cavity with slots for the wall inserts. The frame height sets the thickness of the white layer, $d = 2$~mm; together with the 0.5~mm acrylic base resting on the 0.65~mm sensor packages, the black film, and the 0.2~mm FPCB, the sensing stack is about 3.5~mm thick. Both the FPCB and the acrylic base bend, so the pad can be mounted on curved surfaces; all experiments here use it flat.

\subsubsection{Mode~1 fabrication} The degassed white translucent mixture is poured slowly into the assembled mold to minimize air entrapment as shown in Fig.~\ref{fig:fabrication} (a)(2). An acrylic plate is pressed onto the mold to remove excess silicone and produce a flat layer of uniform thickness (Fig.~\ref{fig:fabrication} (a)(3)). The mold is heat-cured at \SI{65}{\celsius} for approximately 20~min (Fig.~\ref{fig:fabrication}~(a)(4)). The mold frame is then removed, leaving the white layer bonded to the acrylic base. The black mixture is poured over the white layer and allowed to spread across the acrylic, covering the
top and sides of the white translucent layer (Fig.~\ref{fig:fabrication} (a)(5)), and is heat cured for
approximately 10~min (Fig.~\ref{fig:fabrication} (a)(6)). The finished Mode~1 stack is shown in Fig.~\ref{fig:fabrication} (a)(7).

\subsubsection{Mode~2 fabrication} Before casting, a grid-like dummy wall insert is assembled in the mold
where three straight column walls are placed in the frame slots, and three row walls with 1.25~mm half-lap notches are lowered on top so that the two sets interlock and hold each other in position (Fig.~\ref{fig:fabrication} (b)(2)). The white translucent mixture is poured around the walls, leveled, and heat-cured at \SI{65}{\celsius} for approximately 20~min.

Removing the grid leaves one translucent cube per taxel separated by 1~mm slits as in Fig.~\ref{fig:fabrication} (b)(4). The prepared pigmented mixtures (green, red, yellow, and blue) are injected into the slits so that the slits between neighboring columns alternate green/red and the slits between neighboring rows alternate yellow/blue as shown in Fig.~\ref{fig:fabrication} (b)(5). The rows are filled and heat cured first for approximately 10~min, followed by the columns for 10~min (Fig.~\ref{fig:fabrication} (b)(6)). The frame is then removed, and the black coating is applied (Fig.~\ref{fig:fabrication} (b)(7)) and heat cured for 10~min (Fig.~\ref{fig:fabrication} (b)(8)), similar to Mode~1.

In both variants, the finished stack is aligned over the photodiode array on the flexible PCB and fixed in place (Fig.~\ref{fig:fabrication} (a)(7) and (b)(9)).

\section{Characterization}\label{sec:protocol}\label{sec:results}

\subsection{Setup and Normalization}
\sensor is mounted on a rigid base under a three-axis GRBL CNC gantry. A six-axis ATI Gamma force/torque sensor sits between the spindle and the indenter, so it measures the force applied by the indenter regardless of how \sensor is mounted. One script coordinates the gantry, the force/torque stream, and the \sensor stream and time-stamps every sample. The force/torque sensor is tared over \SI{1}{\second} in free air, and the run aborts if the unloaded force at any retracted position deviates from this tare by more than \SI{0.15}{\newton}. The LEDs are switched on 5\,min before each experiment so that LED output and detector dark current settle.

From unloaded frames we obtain each taxel's baseline $b_i$ and noise $\sigma_i$, and from full compression its floor $\raw_i^{\min}$. Because the reading falls with load, we define the normalized signal as:
\begin{equation}
  \nsig_i = \operatorname{clip}\!\left(\frac{b_i - \raw_i}{b_i - \raw_i^{\min}},\,0,\,1\right),\qquad \nsig_i \leftarrow 0 \ \text{if}\ b_i - \raw_i < 3\sigma_i,
  \label{eq:norm}
\end{equation}
so that $\nsig_i = 0$ at no contact and $\nsig_i = 1$ at full compression, with a per-taxel noise floor at $3\sigma_i$. This two-point normalization corrects offset and gain differences between taxels, and the same rule is applied in simulation (Sec.~\ref{sec:sim}).

\subsection{Normal Cartridge}
\textit{Protocol.} A flat circular indenter of \SI{4.5}{\milli\meter} diameter, roughly one taxel in size, is positioned over each taxel center in serpentine order and lowered in \SI{0.05}{\milli\meter} steps at \SI{30}{\milli\meter\per\minute}. After each step it dwells \SI{2}{\second} so that the viscoelastic silicone settles, and we average the last \SI{0.5}{\second} of force, depth, and all taxel readings. Loading stops at \SI{2.5}{\milli\meter} or \SI{6}{\newton}, and the indenter then retracts in the same steps to \SI{0.5}{\milli\meter} above the surface. The full-pad pass is repeated 15 times in five sessions (240 presses). Because LED output drifts slowly, $b_i$ is re-measured for every press from the unloaded frames just before it, and $\raw_i^{\min}$ is taken at the peak load of the taxel's own presses ($\approx$\SI{6}{\newton}), leave-one-pass-out so that no pass normalizes itself. All taxels are recorded on every press, which gives the spatial response and the crosstalk directly.

\begin{figure*}[!tp]
  \centering
  \includegraphics[width=\textwidth]{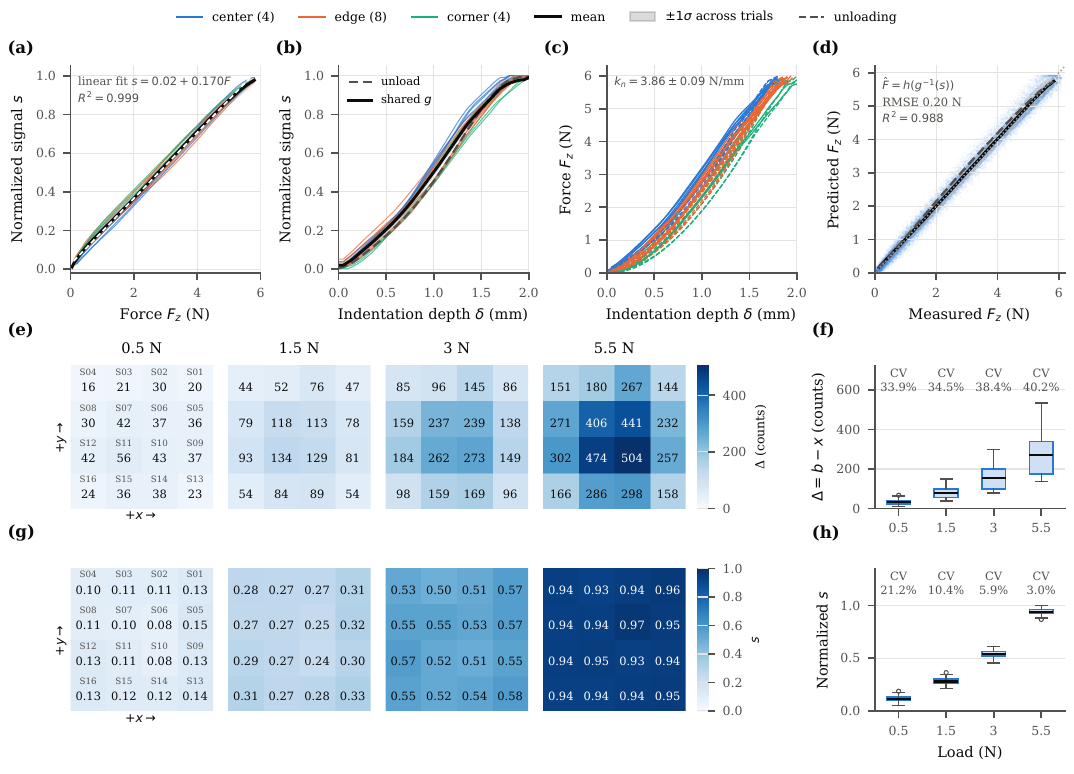}
  \caption{\textbf{Normal cartridge, 240 presses.} (a)~Normalized signal vs.\ force (thin: per-taxel means; black: mean over taxels; band: $\pm1\sigma$ across passes); $\nsig = 0.02 + 0.170F$, $R^2 = 0.999$. (b)~Signal vs.\ depth $g(\dep)$: per-taxel isotonic fits and the shared $g$; dashed: unloading. (c)~Force vs.\ depth, loading (solid) and unloading (dashed). (d)~Force predicted from the signal, $\hat F = h(g^{-1}(\nsig))$, vs.\ measured force, leave-one-session-out (black: binned loading mean, $\pm1\sigma$; dashed: unloading; dotted: ideal). (e)--(h)~Uniformity at 0.5, 1.5, 3, and \SI{5.5}{\newton}: raw drop $\Delta = b - \raw$ of the pressed taxel (e) and its distribution with CV (f); normalized signal (g) and its distribution (h). Maps share one color scale per row and are drawn in the CNC frame.}
  \label{fig:char}
\end{figure*}

\textit{Response.} Raw baselines differ strongly across the array: $2364 \pm 284$ counts at the center, $1322 \pm 203$ at the edges, and $673 \pm 100$ at the corners, which collect scattered light from fewer LEDs. Every taxel nevertheless responds monotonically from the first loading step, with a detection threshold (smallest force with $b_i - \raw_i > 3\sigma_i$) of $104 \pm 49$\,mN, limited in part by the step size. After normalization the response is linear in force (Fig.~\ref{fig:char} (a)): the mean over taxels follows $\nsig = 0.02 + 0.170\,F$ with $R^2 = 0.999$ from \SI{0.1}{\newton} to the \SI{5.8}{\newton} test limit, and no taxel's local sensitivity $\mathrm{d}\nsig_i/\mathrm{d}F$ falls below half its median, so neither a dead zone nor saturation is reached. Hysteresis, the largest loading--unloading difference in $\nsig$ at equal depth, is $5.9 \pm 1.1$\% of full scale, and repeatability, the standard deviation of $\nsig$ at fixed depth across passes, is $2.0 \pm 0.3$\% (Table~\ref{tab:metrics}).

\textit{Uniformity.} We compare the pressed taxel's signal across the array at 0.5, 1.5, 3, and \SI{5.5}{\newton} (Fig.~\ref{fig:char} (e--h)). In raw counts the taxels differ strongly: at \SI{3}{\newton} the drop is 85 counts at a corner but 273 at the center, because center taxels receive more LED light, and the coefficient of variation (CV) across taxels is 34--40\% at every load. After normalization~\eqref{eq:norm} the CV falls to 21\%, 10\%, 5.9\%, and 3.0\%: center, edge, and corner taxels become indistinguishable, and the four loads no longer overlap. The CV at \SI{5.5}{\newton} is partly set by the normalization ($\nsig = 1$ at peak load) and the CV at \SI{0.5}{\newton} by the small signal there, so 1.5 and \SI{3}{\newton} are the representative loads.

\textit{Crosstalk.} We define the crosstalk of neighbor $j$ as $c_{ij} = \nsig_j / \nsig_i$. At light load the neighbors are essentially unchanged ($\nsig \leq 0.03$ at \SI{0.5}{\newton}). At peak load, $-x$ neighbors respond with $0.7 \pm 1.2$\% of the pressed signal, $\pm y$ neighbors with $5.2 \pm 6.0$\%, and diagonal neighbors with $0.8 \pm 1.6$\%. The $+x$ neighbors respond far more, with a median of 30\%, 38\%, and 70\% for presses in the first three columns. It is the same in all five sessions (37--38\%) and passes, so it does not drift, and it is absent in $-x$, whereas optical or mechanical coupling would be symmetric; we therefore attribute it to a fixed indenter-position error that grows along $x$, which may also add to the residual taxel-to-taxel variation.

\textit{Depth response.}\label{sec:calib} Because the reading follows indentation depth, we also fit the normalized signal as a function of depth, $\nsig_i = g_i(\dep)$, per taxel with a monotone (isotonic) regression on the loading branch, and the force as a function of depth, $F = h(\dep)$ (Fig.~\ref{fig:char} (b,c)). The force--depth curves are nearly identical across the array: a linear fit between 20 and 80\% of peak force, divided by the number $N_c$ of taxels under the indenter, gives a per-taxel stiffness $k_n = N_c^{-1}\,\mathrm{d}F/\mathrm{d}\dep = 3.86 \pm 0.09$\,N/mm, and \SI{6}{\newton} is reached at $1.81 \pm 0.07$\,mm. A single shared $g$ reproduces the individual taxels with an RMSE of $0.038 \pm 0.011$. Predicting force as $h(g^{-1}(\nsig))$, with both curves fit on four sessions and tested on the held-out fifth (leave-one-session-out), gives an RMSE of \SI{0.20}{\newton} ($R^2 = 0.988$) (Fig.~\ref{fig:char} (d)).

\begin{table}[t]
\caption{Characterization summary of the two cartridges. Normal cartridge: mean $\pm$ std over taxels. Shear cartridge: held-out data.}
\label{tab:metrics}
\centering
\footnotesize
\setlength{\tabcolsep}{4pt}
\begin{tabular*}{\columnwidth}{@{}l@{\extracolsep{\fill}}c@{}}
\toprule
\multicolumn{2}{@{}l}{\textbf{Normal cartridge}} \\
\midrule
\quad Noise $\sigma$ (counts; \% of baseline) & $1.06 \pm 0.44$; 0.08 \\
\quad Detection threshold (mN) & $104 \pm 49$ \\
\quad Working range (N) & 0.10--5.8$^{a}$ \\
\quad Saturation force (N) & $>5.8$ (not reached)$^{a}$ \\
\quad Sensitivity (FS/N), linearity $R^2$ & $0.170$, $0.999$ \\
\quad Hysteresis (\% FS) & $5.9 \pm 1.1$ \\
\quad Repeatability (\% FS) & $2.0 \pm 0.3$ \\
\quad Crosstalk, $-x$ / $\pm y$ neighbor (\%)$^{b}$ & $0.7 \pm 1.2$ / $5.2 \pm 6.0$ \\
\quad Crosstalk, diagonal neighbor (\%) & $0.8 \pm 1.6$ \\
\quad Per-taxel stiffness $k_n$ (N/mm) & $3.86 \pm 0.09$ \\
\quad Force from $h \circ g^{-1}$, RMSE (N) & $0.20$ ($R^2 = 0.988$) \\
\midrule
\multicolumn{2}{@{}l}{\textbf{Shear cartridge}$^{c}$} \\
\midrule
\quad Force $R^2$, $x$ / $y$ / $z$, unseen positions & $0.85$ / $0.75$ / $0.90$ \\
\quad Force $R^2$, $x$ / $y$ / $z$, held-out session$^{d}$ & $0.96$ / $0.96$ / $0.92$ \\
\quad Per-taxel force $R^2$, held-out session$^{d,e}$ & $0.92$ / $0.93$ / $0.94$ \\
\quad Shear direction error, unseen pos.\ (circle) / session & \SI{14}{\degree} / \SI{4.8}{\degree} \\
\quad Contact location error, unseen positions (mm) & $0.56$ \\
\quad Force repeatability between sessions (N) & 0.05--0.10 \\
\quad Normal linearity $R^2$; noise (\% of baseline) & $0.996$; $0.10$ \\
\bottomrule
\end{tabular*}\\[2pt]
\parbox{0.95\linewidth}{\scriptsize $^{a}$Upper limit set by the \SI{6}{\newton} test limit. $^{b}$At peak load; the $+x$ neighbor is excluded because of an indenter offset (Sec.~\ref{sec:results}). $^{c}$Sum of the per-taxel estimates vs.\ the measured force; unseen positions: \SI{4.5}{\milli\meter} circle; location from the whole-pad decoder. $^{d}$Leave-one-out over nine sessions (plate: three repeats in one sitting), three indenters. $^{e}$Vs.\ labels assuming uniform traction.}
\end{table}
\subsection{Shear Cartridge}\label{sec:protocol_shear}\label{sec:shear_results}
\textit{Protocol.} Three flat indenters are used. A \SI{4.5}{\milli\meter} circle, about one taxel in size, visits 57 positions: 49 on a $7{\times}7$ grid at half the taxel pitch around the pad center, and eight random off-grid test positions that are never used for training. At each position it is lowered until it rests lightly on the gel, where the force/torque reading is zeroed, pressed in \SI{0.2}{\milli\meter} steps (\SI{0.35}{\second} dwell) to \SI{2}{\milli\meter} or \SI{6}{\newton}, and, at normal loads of 2 and \SI{4}{\newton}, moved laterally in four directions out to \SI{2}{\milli\meter} and back in \SI{0.4}{\milli\meter} steps. The sequence was repeated in three sessions (15{,}522 hold steps). A \SI{9}{\milli\meter} square repeats it at 33 positions (a $5{\times}5$ half-pitch grid and the same test positions) with shear at up to \SI{8}{\newton} in three sessions, the last of which presses positions between the trained grid points. A $12{\times}\SI{12}{\milli\meter}$ plate visits nine positions in eight directions at four depths, three times in one sitting (864 sweeps). The rotation between the force/torque and pad frames is fitted in each session from the shear sweeps.

\textit{Features and models.} For taxel $i$ we use the clear-channel brightness change $p_i = (b_i^{C} - C_i)/b_i^{C}$ and the chromaticity changes $\Delta r_i$, $\Delta g_i$, $\Delta b_i$, with $r_i = R_i/C_i$ and likewise for green and blue, relative to their unloaded values. A per-taxel model, one two-layer perceptron (64 units per layer) per taxel on the features of its $3{\times}3$ neighborhood, outputs the three-axis force at that taxel. Because the force/torque sensor measures only the total force, its labels assign the measured force to each taxel in proportion to the contact area over its cell, and its estimates are evaluated by summing them over the taxels and comparing with the measured total, which does not depend on this assumption. A whole-pad kernel ridge regression estimates the total force and the contact location. Splits are always by position or by session.

\begin{figure*}[!t]
  \centering
  \includegraphics[width=\textwidth]{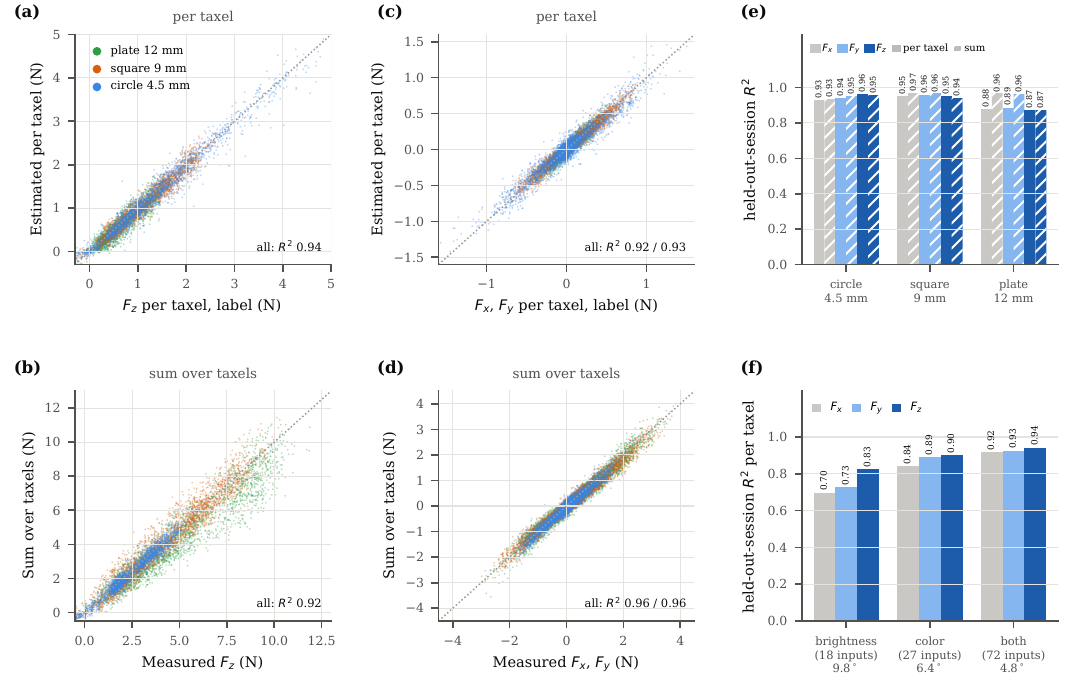}
  \caption{\textbf{Shear cartridge, one model per taxel trained on all three indenters and tested on held-out sessions (each of nine sessions left out in turn).} (a), (c)~Per-taxel normal and lateral force vs.\ the label, the measured force times the share of the contact area over the taxel, for covered taxels (at least 25\% of the contact; plate: cell at least half covered). (b), (d)~Sum of the per-taxel estimates vs.\ the force/torque sensor, which does not depend on the label. (e)~$R^2$ per indenter, per taxel (solid) and summed (hatched). (f)~The same model retrained with brightness only, color only, or both (per-taxel $R^2$, all indenters; median shear-direction error below each group).}
  \label{fig:char_shear}
\end{figure*}

\textit{Three-axis force.} At the eight positions never seen in training, each covered taxel's estimate matches its label with $R^2 = 0.83$, $0.71$, and $0.89$ in $x$, $y$, and $z$ for the circle and $0.94$, $0.93$, and $0.93$ for the square. Summed over the taxels, the estimates reproduce the measured total force with $R^2 = 0.85$, $0.75$, and $0.90$ for the circle and $0.96$ for the square, and the shear direction with a median error of \SI{14}{\degree} and \SI{5}{\degree}. To test new sessions, we hold out each of the nine sessions in turn (three per indenter; for the plate, the three repeats recorded in one sitting) and train on the other eight, with all three indenters together. Each covered taxel then reaches $R^2 = 0.92$, $0.93$, and $0.94$ (RMSE 0.06, 0.05, and \SI{0.14}{\newton}; Fig.~\ref{fig:char_shear} (a, c)), the summed estimates $0.96$, $0.96$, and $0.92$ (Fig.~\ref{fig:char_shear} (b, d)), and the median direction error is \SI{4.8}{\degree}. Accuracy is similar for the three indenters (Fig.~\ref{fig:char_shear} (e)), so each taxel reports its own normal and shear force regardless of contact size. Shear stays at $R^2 \geq 0.93$ in every held-out session. The whole-pad decoder locates the contact at unseen positions with a median error of \SI{0.56}{\milli\meter}, about one-eighth of the pitch. At the same position and step, the measured force differs between sessions by a standard deviation of 0.05--\SI{0.10}{\newton} ($r \geq 0.998$).

\textit{Color carries shear.} To isolate the channels, the same model is retrained with brightness features only ($p$ of the $3{\times}3$ neighborhood) or color features only ($\Delta r$, $\Delta g$, $\Delta b$, with no brightness input). Per taxel, brightness alone reaches $R^2 = 0.70$, $0.73$, and $0.83$ with a median direction error of \SI{9.8}{\degree}, color alone $0.84$, $0.89$, and $0.90$ with \SI{6.4}{\degree}, and both together $0.92$, $0.93$, and $0.94$ with \SI{4.8}{\degree} (Fig.~\ref{fig:char_shear} (f)). Color thus carries most of the shear, brightness adds to the normal force, and the two are complementary. Consistent with the wall layout, the red--green change of each taxel follows $F_x$ and the yellow--blue change follows $F_y$, with signs alternating between neighboring columns and rows.

\textit{Contact shape.} Because \sensor measures deformation, the force estimate depends on contact size. Retraining the model on every combination of the three indenters shows that contacts larger than one taxel transfer to each other (normal force $R^2 = 0.82$ from square to plate and $0.80$ from plate to square), but none transfers to the one-taxel circle ($R^2 \leq 0.18$), so the training set should include the smallest expected contact. With the circle and the square in training, the summed estimate recovers the plate, which is never used for training, with $R^2 = 0.86$, $0.94$, and $0.85$, and a single model trained on all three matches the best result on every indenter, so one model serves all contact sizes.

\textit{Normal sensing.} The shear cartridge still senses pressure linearly ($R^2 = 0.996$ against load for taxels at least half covered by the plate), with noise similar to the normal cartridge (0.10\% vs.\ 0.08\% of baseline).

\textit{Shared model.} The models above are trained per taxel, from about 60 labeled press-and-shear sequences over each taxel. To test whether larger arrays can avoid this, we trained a single model shared by all taxels on three quadrants of the pad and adapted it to each taxel of the fourth with $K$ labeled sequences from that taxel ($K$-shot). Without adaptation the model does not transfer (summed $R^2 = 0.33/0.47/0.55$); with $K = 1$, $3$, and $10$ it reaches $0.77/0.75/0.72$, $0.83/0.81/0.71$, and $0.88/0.86/0.79$, approaching per-taxel training ($0.96/0.96/0.92$) with about one sixth of its data, whereas a per-taxel model trained on the same ten sequences reaches only $0.74/0.75/0.70$.

\begin{figure}[!t]
  \centering
  \includegraphics[width=\linewidth]{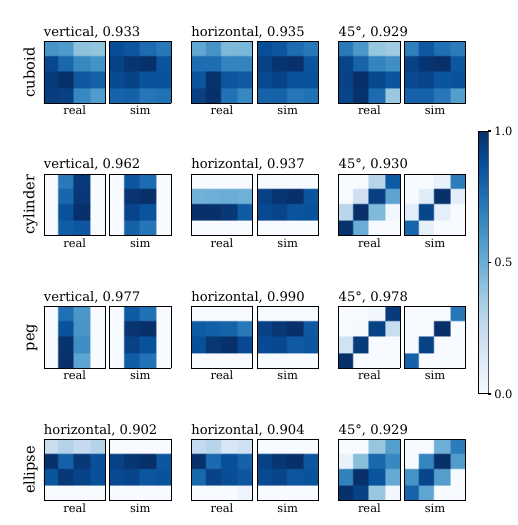}
  \caption{\textbf{Measured and simulated readings of \sensor for four objects in three
  orientations.} Each grid is the mean of the 50 frames of one hold, normalized by its
  peak; the number is the mean cosine over all frame pairs of that grasp.}
  \label{fig:sim2real}
\end{figure}

\section{Simulation}\label{sec:sim}
We simulate \sensor so that tactile data can be collected and policies trained in simulation.

\subsection{Simulated Sensor}\label{sec:sim_impl}
We simulate \sensor in IsaacGym~\cite{makoviychuk2021isaac} with the tactile simulation of TacSL~\cite{akinola2024tacsl}. The simulated pad is a gel block on the finger of a parallel-jaw gripper mounted on a Franka Panda arm, as in the real grasps. Each taxel is one sample point on the outer surface of the gel, above its detector. At every step the simulator reads the indentation depth $\dep_i$ of each point from the object's signed distance field.

Section~\ref{sec:calib} shows that a taxel's reading is set by how far the gel above it is pressed in, $\nsig_i = g_i(\dep_i)$, and rises monotonically with depth. The simulator applies the same $g_i$ to the simulated depth. Because $g_i$ is monotonic, the real reading also gives depth, $\dep_i = g_i^{-1}(\nsig_i)$, so the real and simulated pads report the same quantity: the local indentation of the gel.

For simulation we write $g_i$ in closed form. Let $D_i = (b_i - \raw_i)/b_i$ be the fractional drop of taxel $i$ below its baseline. On the presses of Sec.~\ref{sec:calib} we fit it per taxel as $D_i = \alpha_i\,\dep^{\,\beta_i}$. Dividing numerator and denominator of \eqref{eq:norm} by $b_i$ gives the same rule in these units, $\nsig_i = \operatorname{clip}(D_i/D_i^{\max},0,1)$, where $D_i^{\max} = (b_i - \raw_i^{\min})/b_i$ is the drop at full compression (about \SI{6}{\newton}). The simulated signal is
\begin{equation}
  \hat{\nsig}_i = \operatorname{clip}\!\left(\frac{\alpha_i\,\dep_i^{\,\beta_i}}{D_i^{\max}},\,0,\,1\right),
  \label{eq:simsignal}
\end{equation}
with $\hat{\nsig}_i = 0$ when $\alpha_i\,\dep_i^{\,\beta_i} < 3\sigma_i/b_i$, the noise gate of \eqref{eq:norm} in the same units. The simulator needs four numbers per taxel from the real pad, $\alpha_i$, $\beta_i$, $D_i^{\max}$, and $\sigma_i/b_i$, and no raw counts, baselines, or LED drift. Since crosstalk between neighboring taxels is small (Sec.~\ref{sec:calib}), no spatial blur is applied. The simulator will be released with the \sensor repository.

\begin{figure*}[!t]
    \centering
    \includegraphics[width=\textwidth]{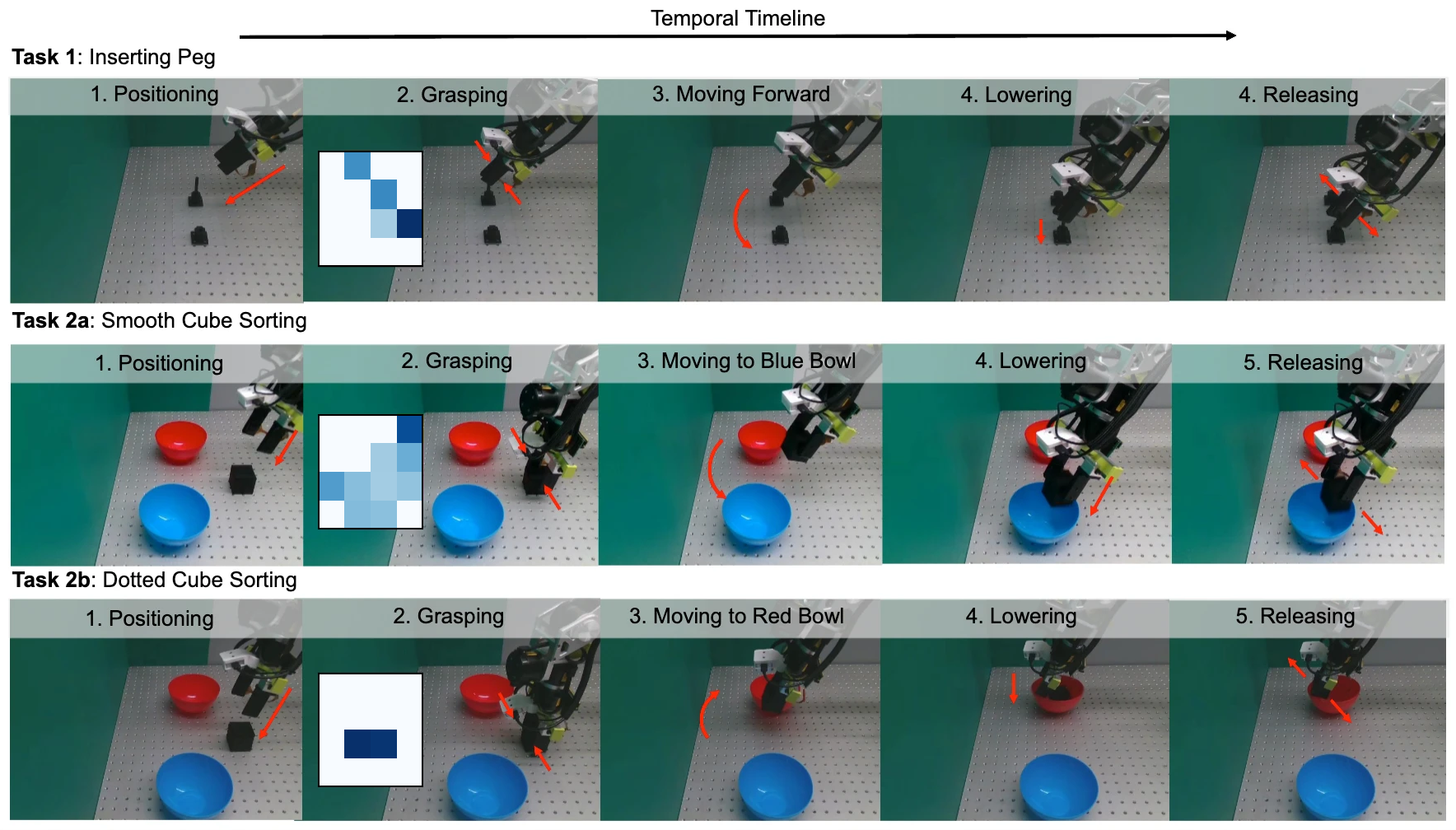}
    \caption{\textbf{
    Temporal tactile observations during end-to-end manipulation using \sensor.}
    Representative execution sequences for (Task 1) peg-in-hole insertion and (Task 2) cube sorting with smooth and dotted surfaces. Each row illustrates successive manipulation stages, including positioning, grasping, object transport, lowering, and release. The corresponding LuxTac heatmaps reveal contact-dependent responses during manipulation.
    }
    \label{fig:temporal_observations}
\end{figure*}

\begin{table}[!t]
\caption{Sim-to-real alignment of the normal cartridge (12 grasps $\times$ 50 frames per
side). Rows 1--7 are mean cosines between the 600 real frames and each comparison.}
\label{tab:sim2real}
\centering
\footnotesize
\begin{tabular*}{\columnwidth}{@{}l@{\extracolsep{\fill}}c@{}}
\toprule
Real frames compared against & Score \\
\midrule
Real frames of the same hold (ceiling) & 0.9994 \\
\textbf{Matched simulation} & \textbf{0.942} \\
\midrule
Uniform pad & 0.763 \\
Simulation of a different object & 0.674 \\
Same object, different orientation & 0.642 \\
Matched simulation, taxels shuffled & 0.594 \\
\midrule
\SI{45}{\degree} grasps: correct / mirrored diagonal & 0.946 / 0.264 \\
One-to-one assignment, object recovered & 10 / 12 \\
Wasserstein-1 distance of pooled values & 0.067 \\
\bottomrule
\end{tabular*}
\end{table}

\subsection{Alignment Protocol and Results}\label{sec:sim2real}
Four 3D-printed prisms, a $20{\times}\SI{20}{\milli\meter}$ cuboid, a \SI{25}{\milli\meter} cylinder, a $40{\times}\SI{20}{\milli\meter}$ ellipse, and a \SI{15}{\milli\meter} peg, were each grasped three times with the normal cartridge, in vertical, horizontal, and \SI{45}{\degree} orientations, and 50 frames of each static hold were recorded (12 grasps, 600 frames). The same 12 grasps were run in simulation with CAD geometry, with 50 frames each.

\textit{Score.} Real and simulated frames go through the same two steps. First, each taxel is normalized to $[0,1]$: the real taxels by \eqref{eq:norm} and the simulated taxels by its equivalent \eqref{eq:simsignal}, with the same full-scale value and noise gate. Second, every frame is divided by its largest taxel value, so that the score compares the contact pattern. The similarity of two frames is the cosine between their 16 taxel values. A static hold has no frame-to-frame correspondence, so each real frame is scored against every simulated frame of the same grasp and the scores are averaged per grasp. Averaged over the 12 grasps, the cosine between real and simulated frames is 0.942 (median 0.934, lowest grasp 0.902; Fig.~\ref{fig:sim2real}).

\textit{References.} A cosine between non-negative vectors does not start at zero, so Table~\ref{tab:sim2real} places 0.942 between the repeatability of the real sensor (two real frames of the same hold agree at 0.9994) and deliberately wrong predictions: a uniform pad, the simulation of a different object or orientation, and the matched simulation with shuffled taxels score 0.594--0.763. A high score therefore requires the right object, in the right orientation, with the values at the right taxels.

\textit{Further checks.} For the cylinder, ellipse, and peg at \SI{45}{\degree}, simulating the grasp along the opposite diagonal drops the score from 0.946 to 0.264, below the uniform pad. Assigning each real grasp to a distinct simulated grasp (Hungarian assignment) recovers the object for 10 of 12 grasps; the two misses, the horizontal ellipse and peg, both leave a single band across the pad. Finally, the normalized taxel values pooled over all frames differ between real and simulation by a Wasserstein-1 distance of 0.067 on $[0,1]$, about 7\% of full scale.

\section{Robot Experiments}\label{sec:robot}

\subsection{Policy Formulation}
We learn end-to-end visuotactile policies from human demonstrations, without explicit force estimation or handcrafted contact models, using the Action Chunking Transformer (ACT)~\cite{zhao2023act}. The policy conditions on $O_t = \{I_t^{\mathrm{ext}}, \mathbf{s}_t^{\mathrm{tactile}}, \mathbf{p}_t\}$: the external RGB image, which gives the scene; the \sensor observation, the 16 taxel readings as a flat vector, which gives the local contact; and the robot state. It predicts a chunk of future actions $\mathbf{a}_{t:t+H-1} \in \mathbb{R}^{H\times d_a}$ over a horizon $H$.

\subsection{Experimental Platform and Tasks}
Experiments were conducted on a reBot Arm B601 with a custom 3D-printed gripper that carries one \sensor pad on one finger, and a side-facing RGB camera. The policy receives the per-taxel pressure signal of \sensor; shear is not yet used as a policy input. We consider two tasks.

\textbf{Peg-in-hole:} the robot grasps a peg, transports it to the insertion location, lowers it into the hole, and releases it, which requires accurate positioning and controlled contact.

\textbf{Cube sorting:} the robot sorts smooth cubes into a blue bowl and dotted cubes into a red bowl, which tests whether touch helps surface discrimination.

For each task, 50 human demonstrations with synchronized multimodal observations and robot actions were collected on this platform. Both policies use the default ACT implementation of LeRobot~\cite{cadene2024lerobot}. We train two ACT policies: a vision-only baseline using the external RGB image and robot state, and a visuotactile policy that additionally receives \sensor measurements, which isolates the contribution of touch.

\subsection{Results}
Success rates are reported with 95\% Wilson score intervals, and the two policies are compared with Fisher's exact test. Fig.~\ref{fig:temporal_observations} shows representative executions; the \sensor heatmaps show distinct contact responses during grasping that complement the external view.

For peg-in-hole insertion, \sensor raised success from 6 to 13 of 15 trials, a significant improvement (Fisher's exact test, $p = 0.021$; Table~\ref{tab:task_success_rates}). For cube sorting, success changed from 5 to 6 of 15, which is not significant ($p = 1.0$); distinguishing smooth from dotted cubes requires resolving surface features that may be finer than the 4.4--\SI{4.8}{\milli\meter} taxel pitch of the current prototype. Tactile feedback from \sensor therefore helps most where contact governs success, as in precision insertion.

\begin{table}[!t]
\centering
\caption{Real-world success rates for peg-in-hole and cube sorting
tasks, with 95\% Wilson confidence intervals in brackets.}
\label{tab:task_success_rates}
\renewcommand{\arraystretch}{1.2}
\setlength{\tabcolsep}{4pt}
\resizebox{\columnwidth}{!}{%
\begin{tabular}{lccc}
\toprule
\textbf{Method} &
\textbf{Peg-in-Hole} &
\textbf{Cube Sorting} &
\textbf{Avg.} \\
\midrule

Vision Only
& 6/15 (40\%) [20--64]
& 5/15 (33\%) [15--58]
& 37\% \\

\textbf{Vision + LuxTac}
& \textbf{13/15 (87\%)} [62--96]
& \textbf{6/15 (40\%)} [20--64]
& \textbf{63\%} \\

\bottomrule
\end{tabular}%
}
\end{table}

\section{Discussion and Future Work}\label{sec:discussion}
Because \sensor measures deformation, its force estimate depends on contact shape: a sharp object indents further than a flat one at the same force, and a model trained only on contacts larger than one taxel fails on the one-taxel circle (Sec.~\ref{sec:shear_results}). Training on several sizes, including the smallest, closes this gap; for learning, the dependence is also a feature, since the policy receives the same geometric quantity in simulation and reality, whereas force control would need a contact model such as Hertzian contact~\cite{johnson1985contact}. Because the reading follows deformation, pairing it with force also reveals how compliant the object is, which learned tactile sensing has used to estimate Young's modulus and Poisson's ratio~\cite{xie2026synthetic}. The silicone also creeps under sustained load, and our stepped protocol characterizes the quasi-static response. LED output drifts with temperature, so the baseline is re-estimated at the start of each episode. The shear cartridge adds a three-axis force estimate at every taxel without new electronics; it has been evaluated with three flat indenters at held-out positions and sessions, while curved, sharp, and deformable objects remain to be tested. With the same \SI{4.5}{\milli\meter} indenter, the colored walls of this first shear cartridge lower the clear-channel drop at \SI{1.8}{\milli\meter} from 21.9\% to 15.9\% of baseline and raise hysteresis from 5.9\% to 9.1\% FS, partly because the shear protocol slides the indenter between loading and unloading, while repeatability stays similar (2.7\% vs.\ 2.0\% FS). The normal cartridge remains the default when only pressure is needed.

Because a shared model adapts to a new taxel from a few labeled sequences (Sec.~\ref{sec:shear_results}), larger arrays need not be calibrated from scratch taxel by taxel. On the prototype, most taxels lie on the border, so each held-out taxel has a layout absent from training; arrays with many interior taxels, and indenters with several tips that label many taxels in one press, should narrow the remaining gap.

The simulator reproduces the real contact patterns, while force-controlled grasps, policy training in simulation, and simulation of shear remain to be evaluated. In future work, we will characterize larger arrays with more contact shapes and objects, refine the wall geometry of the shear cartridge, use per-taxel shear as a policy input on the real robot and use the simulator to train policies in simulation, including simulated shear, and transfer them to the real sensor.

\section{Conclusion}
We presented \sensor, a flexible optical taxel array in which each taxel reports the local indentation of a scattering silicone layer under an absorbing film. Because every taxel is an independent digital device, the same cell tiles into larger arrays without changing the readout, and because each detector also reports color, swapping the gel cartridge turns a pressure array into one that resolves three-axis force at every taxel, with no camera, magnets, or bump overlay. The normal cartridge is linear and uniform after a two-point normalization, and its depth response transfers directly into simulation; the shear cartridge estimates normal and shear force per taxel across contact sizes and sessions; and even pressure alone substantially improved a learned insertion policy. Next, we will use per-taxel shear in policies and train them in simulation. We will open-source all hardware and software, including the PCB design, fabrication procedure, firmware, characterization scripts, and simulator integration.

\bibliographystyle{IEEEtran}
\bibliography{refs}

\end{document}